\documentclass[lettersize,journal]{IEEEtran}

\usepackage{amsmath, amsfonts, amssymb, mathtools}
\usepackage[ruled]{algorithm2e}
\usepackage{algorithmic}
\usepackage{array}
\usepackage[caption=false,font=normalsize,labelfont=sf,textfont=sf]{subfig}
\usepackage{tip}
\usepackage{textcomp}
\usepackage{stfloats}
\usepackage{url}
\usepackage{verbatim}
\usepackage{graphicx}
\usepackage{color,xcolor}
\usepackage{hyperref}
\usepackage[capitalize]{cleveref}
\usepackage{bbding}
\usepackage{tabularx}
\usepackage{multirow}
\usepackage{xcolor}

\def\BibTeX{{\rm B\kern-.05em{\sc i\kern-.025em b}\kern-.08em
    T\kern-.1667em\lower.7ex\hbox{E}\kern-.125emX}}

\usepackage{balance}
\begin{document}
\title{Towards Generic and Controllable Attacks Against Object Detection}
\author{Guopeng Li, Yue Xu, Jian Ding, Gui-Song Xia
\thanks{

Guopeng Li is with the School of Computer Science, Wuhan University, Wuhan 430079, China. E-mail: guopengli@whu.edu.cn.

Yue Xu is with the School of Cyber Science and Engineering, Wuhan University, Wuhan 430079, China. E-mail: leoxuy@whu.edu.cn.

Jian Ding is with the State Key Lab. LIESMARS, Wuhan University,
Wuhan, 430079, China. Email: jian.ding@whu.edu.cn.

G.-S. Xia is with the National Engineering Research Center for Multimedia Software, School of Computer Science and Institute of Artificial Intelligence, Wuhan University, Wuhan, 430072, China. Email: guisong.xia@whu.edu.cn.
}}


\maketitle

\begin{abstract}
Existing adversarial attacks against Object Detectors (ODs) suffer from two inherent limitations. Firstly, ODs have complicated meta-structure designs, hence most advanced attacks for ODs concentrate on attacking specific detector-intrinsic structures, which makes it hard for them to work on other detectors and motivates us to design a generic attack against ODs. Secondly, most works against ODs make Adversarial Examples (AEs) by generalizing image-level attacks from classification to detection, which brings redundant computations and perturbations in semantically meaningless areas (\eg, backgrounds) and leads to an emergency for seeking controllable attacks for ODs. To this end, we propose a generic white-box attack, LGP (local perturbations with adaptively global attacks), to blind mainstream object detectors with controllable perturbations. For a detector-agnostic attack, LGP tracks high-quality proposals and optimizes three heterogeneous losses simultaneously. In this way, we can fool the crucial components of ODs with a part of their outputs without the limitations of specific structures. Regarding controllability, we establish an object-wise constraint that exploits  foreground-background separation adaptively to induce the attachment of perturbations to foregrounds. Experimentally, the proposed LGP successfully attacked sixteen state-of-the-art object detectors on MS-COCO and DOTA datasets, with promising imperceptibility and transferability obtained. Codes are publicly released in \url{https://github.com/liguopeng0923/LGP.git}.
\end{abstract}

\begin{IEEEkeywords}
object detection, generic attacks, adversarial examples, controllable imperceptibility.
\end{IEEEkeywords}

\section{Introduction}
\label{sec:intro}

\begin{figure}[!t]
\centering
\includegraphics[width=.9\linewidth]{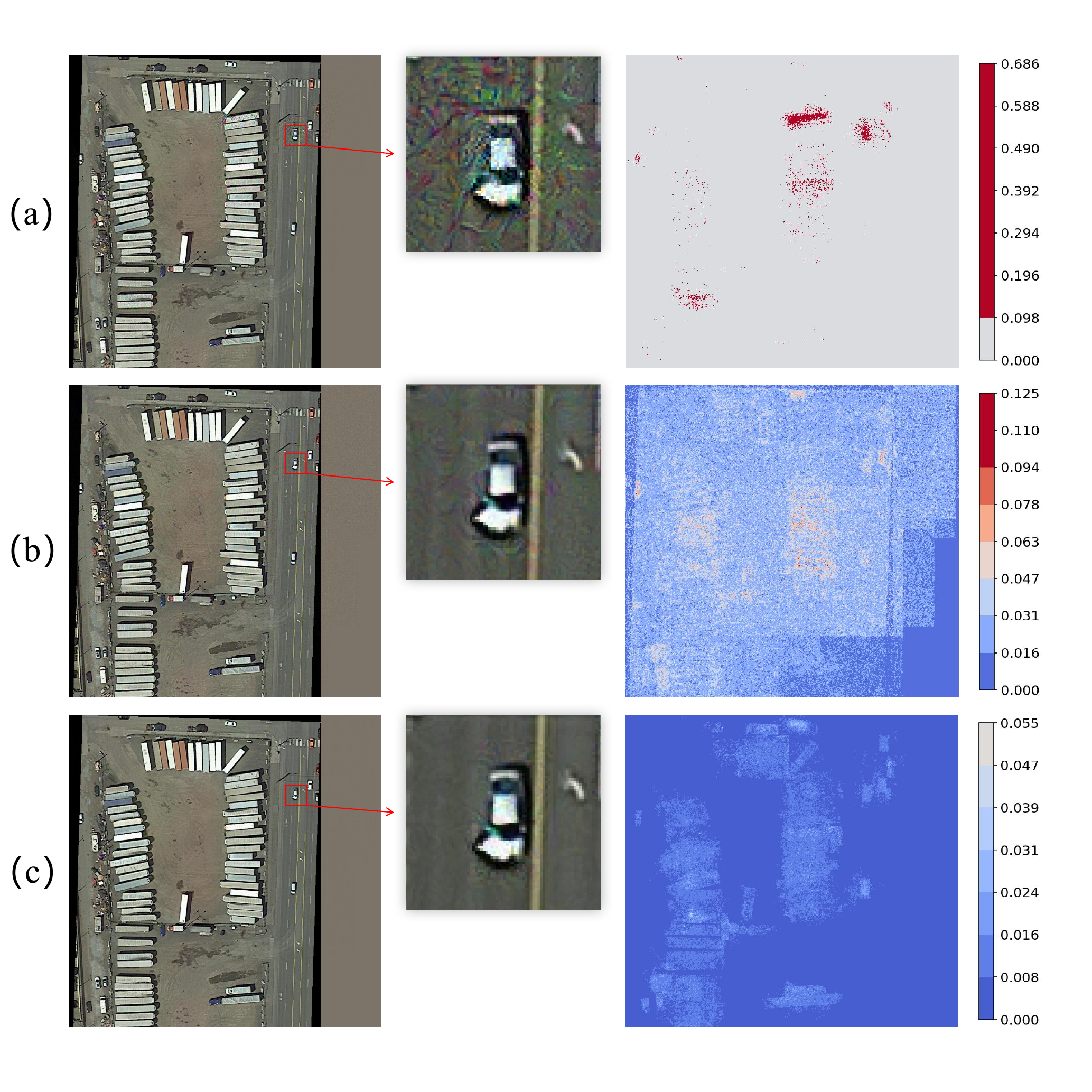}
  \caption{Comparison of Adversarial Examples (left) and perturbations (right) generated by different attack methods: (a) DAG (b) CWA, and (c) our LGP. For visualization, we normalize all perturbations where only \textcolor{blue}{blue} means no perturbations and \textcolor{red}{red} means high perturbations. As shown, adversarial perturbations produced by LGP are with smaller values while mainly attaching to objects.}
  \label{fig:pic/1/per.pdf}
\end{figure}

\IEEEPARstart{I}{mage} understanding technology ~\cite{vggnet,GoogLeNet,resnet,shi2022symmetric} has been dramatically advanced by deep neural networks (DNNs). Nevertheless, they are vulnerable to \emph{adversarial examples} (AEs) with human-imperceptible perturbations and yield erroneous predictions~\cite{szegedy2013intriguing,FGSM}. Such vulnerability inspires increasing attention on the effective attack because it can not only explain the internal mechanism of DNNs to some extent~\cite{ilyas2019adversarial,szegedy2013intriguing} but also help to improve the robustness of learning-based models~\cite{madry2018towards,zhang2019towards,dong2022adversarially}.

As one of the fundamental tasks, object detection has been attracting a lot of attention. The main applications of object detectors (ODs) can be divided into common\cite{faster-rcnn,YOLO,reppoints} and aerial \cite{OrientedRCNN,RoITransformer,S2ANet} object detection and have got impressive accuracy. Even so, there are fewer systemic attacks in the robustness of object detectors compared with the extensive studies in attacking classifiers~\cite{FGSM,PGD,CW,DeepFool,onepixel,xu2020assessing}. An object detector with both high precision and high robustness helps more applications, especially for security sciences such as automatic driving, privacy protection, and \etc. This poses an emergency for studying adversarial attacks on object detection.

For deep object detectors (ODs), the classification networks (\eg, VGG\cite{vggnet}, ResNet\cite{resnet}, \etc) are usually used as feature extractors (\ie, backbone). Besides, ODs contain many other components such as RPN~\cite{faster-rcnn}, ROI Pooling~\cite{faster-rcnn,he2017mask} and prediction heads (\ie, classifying and regressing candidates), as well as non-maximal suppression (NMS)~\cite{NMS} and heuristic label assignment procedures~\cite{zhu2020autoassign}. Different ODs have different structures, thereafter leading to a more complicated problem configuration for the study of adversarial attacks. Generic attacks help us study the roles of different components in the complex pipeline of ODs and inspire a better way for transferable attacks. In this paper, we are interested in studying the challenging problems from the \textbf{universal} nature of ODs. 

\begin{figure}[!t]
  \centering
  \includegraphics[width=0.9\linewidth]{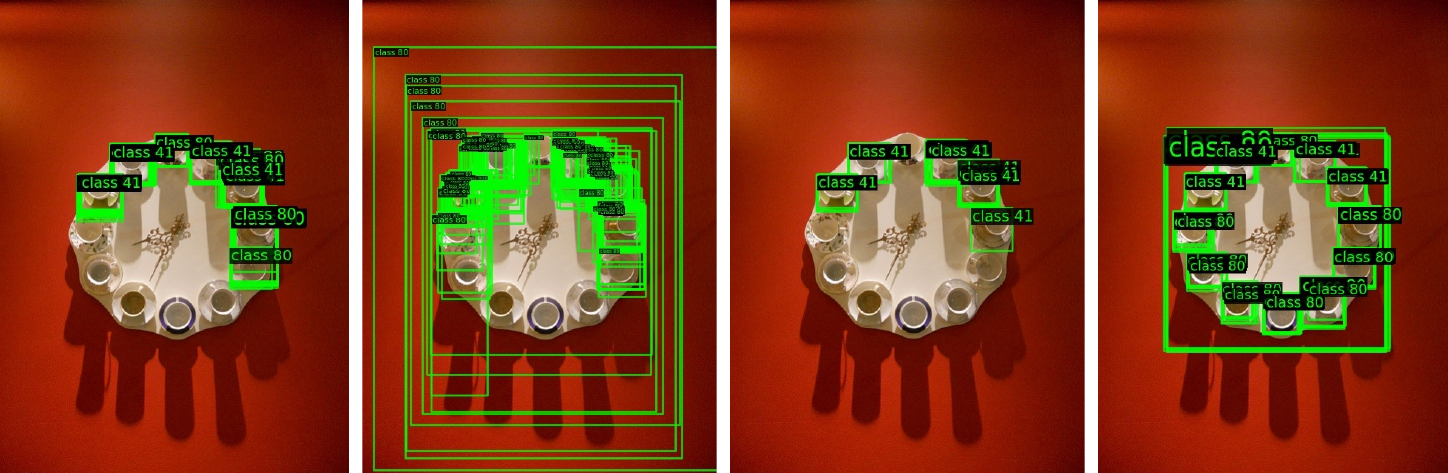}
  \caption{Postive proposals with different methods.
The pictures above are DAG\cite{DAG}, RAP\cite{RAP}, CWA\cite{CWA}, and \textbf{Ours} from left to right. Our Assigner considers adequately high-quality proposals.}
  \label{Fig:pic/1/Assigner.pdf}
\end{figure}

One significant difference between image classification and object detection is the number of candidates. Most highly-performing ODs compute a number of proposals as candidates before post-processing. In this case, existing attacks filter low-quality candidates by modifying some components of ODs (\eg, NMS thresholds of RPN \cite{DAG}, anchor numbers \cite{TOG}, and score thresholds \cite{RAP, CWA}) for generating implicitly image-level proposals awaiting attacks. In other words, existing methods attack some intrinsic structures of ODs, impeding their generalization for new detectors without those specific structures.  Besides, implicit filters inevitably focus on a part of objects with many proposals while omitting the remained objects (\eg, the first three columns in \cref{Fig:pic/1/Assigner.pdf}), making the importance among different objects uncertain. We term such a challenging situation as \textbf{uncertainty of objects attacking}

The multi-task nature of object detection leads to another critical difference. Object detectors usually use multiple prediction branches to learn heterogeneous information about classification likelihoods and object dimensions. Influenced by successes of adversarial attacks for image classification networks, most research approached attacking with a particular focus on classification branch~\cite{DAG,CWA,CA,TOG}. However, AEs generated by optimizing single loss are weakly attacking and have more limitations for improving the robustness of ODs\cite{zhang2019towards,dong2022adversarially}. Although some studies\cite{RAP} attempted to attack multi-task branches jointly, as pointed out in~\cite{zhang2019towards,dong2022adversarially,larget-scale}, forced combinations of misaligned objectives are adverse to joint optimization. We term such a challenge as \textbf{conflicts 
among heterogeneous losses}.

Furthermore, while ``objects" are key to adversarial attacks on object detection, AEs generated by image-level constraints (\eg, clipping\cite{PGD,DAG,CWA} perturbations with $\ell_{p}$ norms\cite{CW}) would inevitably pay more attention to the global context of images instead of objects.  The \textbf{uncontrollable adversarial perturbations} are unsuitable for launching a personalized attack on each object, posing a potential risk of over-perturbation for each object. As shown in \cref{fig:pic/1/per.pdf}(a), the image-level attack will uncontrollably generate easy-to-perceive perturbations on certain objects (\eg, the small car bears many perturbations) and smooth backgrounds. An object-wise attack can generate  personalized perturbations attached to objects without the influence of environments (\eg, (c) in \cref{fig:pic/1/per.pdf}), which is more meaningful and helpful for applications in videos  or reality than an image-level attack against ODs.

In this article, we propose a generic and controllable attacking framework, \ie, local perturbations with adaptive global attacks, named {\em LGP}, which mitigates all challenges above from the universal nature of ODs. In terms of \textbf{uncertainty}, we only attack a small part of ODs'outputs without modifying their inherent structures and explicitly select top-k targets for each object (\eg, the last picture of \cref{Fig:pic/1/Assigner.pdf}). Specifically, we first get the raw outputs of victim models, then assign each object fixed high-quality original proposals based on clean images, and keep track of best matches with them as targets waiting to be attacked based on $i$-th AEs to ensure the entire attacking process is stable (attack relatively fixed targets facing AEs in different iterations). 
For the problem of \textbf{conflicts} among different losses, we set a high-level semantic objective, Hiding Attack (HA)~\cite{jia2022fooling,TOG,yin2022adc}, to guide the entire optimization. In detail, we make a balanced multi-objective loss that simultaneously attacks high-quality candidates from three perspectives: the shape, location, and classification of proposals. In this case, our method consistently minimizes the difference between the distribution of attacked targets and background. 
To improve the \textbf{controllable} imperceptibility of AEs in object detection, \ie, the magnitude, position, and distribution of perturbations, our proposed method adds an adaptive local limit to joint optimization with the attacking objective. As shown in \cref{fig:pic/1/per.pdf} and more results in supplementary materials, LGP focuses on perturbing semantic regions, such as objects in the scene, while suppressing redundant perturbations on irrelevant regions.


The main contributions of this work are threefold. 
\begin{list}{}{}
\item{1) We present a generic (detector- and dataset-agnostic) white-box framework, LGP, against object detection. LGP doesn't need to alter attack strategies and even hyperparameters against new detectors or datasets.}
\item{2) We propose a controllable object-wise constraint to limit the distribution of perturbations adaptively. This is the first insight for controlling the magnitude, position, and distribution of perturbations from ODs' behaviors.}
\item{ 3) Experimental results on sixteen state-of-the-art detectors and two distinct datasets (DOTA~\cite{DOTA} and MS-COCO~\cite{COCO}) demonstrate that our method can yield powerful, controllable, imperceptible, and transferable adversarial perturbations.}
\end{list}

The rest of this article is organized as follows. Section II introduces the related works. Section III states problem definitions and formulations. Section IV describes the details of our method. In Section V, the experimental results and analysis are reported on challenging MS-COCO and DOTA data sets. Finally, the conclusion is made in Section VI.

\section{Related Work}

\textbf{Object Detection}. Object detection aims to localize and recognize objects of interest from images, commonly formulated as a multi-task learning problem. Most detectors can be roughly divided into one-stage\cite{FCOS,reppoints,S2ANet} and two-stage\cite{faster-rcnn,SparseRCNN,OrientedRCNN} detectors. They usually involve feature extraction\cite{resnet}, multi-components (\eg, RPN and RoI\cite{faster-rcnn}) for producing a redundant set of bounding boxes\cite{RetinaNet,GFocalLoss} with classification scores, and performing post-processing such as NMS\cite{NMS} for the final sparse predicts. More recently, End-to-End detectors (\eg, Sparse R-CNN\cite{SparseRCNN}, DETR\cite{DETR}, DiffusionDet\cite{chen2022diffusiondet}, and \etc) produce direct results from learnable sparse proposals without traditional architectures (\eg, anchor, RPN, and NMS). Generally speaking, most ODs have special components, unique architectures, and complex behaviors for final predictions.

\begin{figure*}[!t]
  \centering
  \centering
    \includegraphics[width=.97\linewidth]{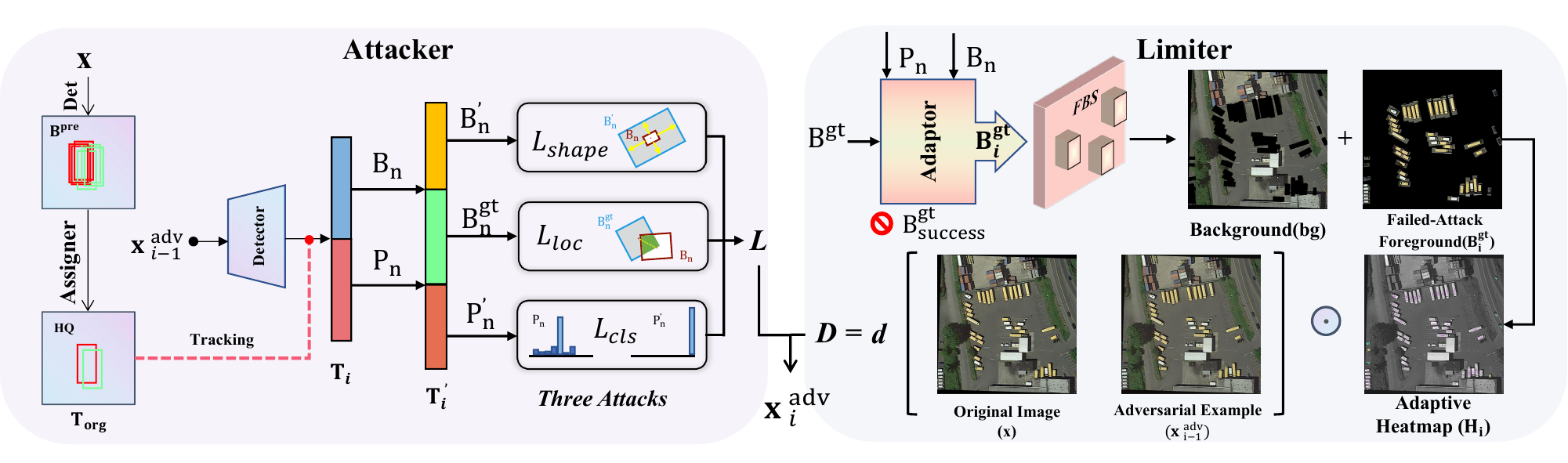}

    \caption{\textbf{The overall pipeline of proposed LGP.} Firstly, we generate fixed original targets ${\cal{T}}_{org}$ from pre-NMS or raw outputs ${\mathbf{B}_{pre}}$ of ODs based on clean images $\mathbf{x}$. Then, we construct targets awaiting attack ${\cal{T}}_{i}$ by matching original targets and pre-NMS outputs based on the last AEs ($\mathbf{x}^{adv}_{i-1}$). Secondly, we set adversarial targets ${\cal{T}}_{i}^{\prime}$ (including bounding boxes ${\cal{B}}_{n}$ and classification probability ${\cal{P}}_{n}$) and combine three different attacking losses $\cal{L}$ for pushing the distribution of ${\cal{T}}_{i}$ to ${\cal{T}}_{i}^{\prime}$ from shape, localization, and semantics. Thirdly, we split the foreground and background of images to compute an adaptive object-wise Heatmap according to the failed-attack foregrounds ($\mathbf{B}_{i}^{gt}$ generated by proposal mappings) for controlling  the distribution of perturbations. Lastly, LGP generates AEs ($\mathbf{x}_{i}^{adv}$) with the joint optimization of attacking $\cal{L}$ and imperceptibility $\cal{D}$ losses.}
    \vspace{-0.2cm}
  \label{fig:pic/overall.pdf}
    
\end{figure*}
\textbf{Adversarial Attack against Object Detectors.} Because of the complex and diverse pipelines of ODs, most existing attacks\cite{DAG,CA,CWA,RAP,wang2021daedalus} focus on specific modules or types of ODs, which impedes their ability to attack new detectors. DAG\cite{DAG} is the first white-box method aiming at RPN-based models\cite{faster-rcnn} by attacking the classifier. Similarly, RAP\cite{RAP} proposes a loss of predicting boxes and classification based on RPN. Besides, CA\cite{CA} and CWA\cite{CWA} take advantage of a weighted class-wise loss for one-stage detectors. Daedalus\cite{wang2021daedalus} creates many false positives by destroying NMS\cite{NMS}. Dpatch\cite{liu2018dpatch} uses visible patches to attack YOLO\cite{YOLO}. Although TOG\cite{TOG} and GAN-based attacks \cite{UEA,aich2022gama} can be viewed as generic attacks, we need to alter their attack strategies making it more complex to apply them to a new problem/dataset. TOG uses RPN to attack RPN-based ODs and anchor-shift to attack anchor-based ODs at that time. But it cannot attack more recent detectors without those basic components (\eg, D.DETR [\textcolor{green}{70}], Sparse R-CNN [\textcolor{green}{44}], and \etc). UEA and GAMA are GAN-based methods, \cite{CA,he2022transferable} points they need to be retrained for launching a new detector/dataset attack leading to more time and data cost than the optimization-based method. UEA also uses RPN loss which limits its attack strength for RPN-free ODs. Moreover, GAN-based methods have more transferable but poorer white-box ability than other optimization-based models. In this paper, we use a unified strategy to attack a small part of ODs' raw outputs without limitations of specific ODs' structures, which induces a generic optimization-based attack.


\textbf{Imperceptible Attack.} Adversaries often need to trade between attack strength and imperceptibility of perturbations, which inspires a lot of works \cite{PGD,CW,frequency,FID} to find a reasonable constraint for evaluating the imperceptibility. However, current attacks for ODs\cite{DAG,RAP,CA,CWA,TOG} clip perturbations based on image level (\ie, they only control the max magnitude of perturbations), which indicates potential uncontrollability (\ie, the position and distribution of learned perturbations is random\cite{frequency,Sharif_2018_CVPR_Workshops}).  To circumvent this problem, \cite{fan2020sparse} factorizes perturbations into magnitude and position vectors, and \cite{frequency} limits perturbations in frequency space from the global image-based viewpoint against image classifiers. They leverage implicitly the models' attention to guide the perturbations, which brings a big learning burden for neural networks and produces suboptimal results.  Differently, we control the magnitude, position, and distribution according to direct proposal mappings and decompose images into foreground-background pairs with an adaptive \textbf{object-wise} constraint motivated by the ``object-centered" behaviors of ODs. In this way, we can launch a local attack for each object while ensuring the global precision drop.

\section{Problem Statement}

An object detector ${\mathcal{D}}et(\mathbf{x})$ takes an input clean image $\mathbf{x}$ as input and outputs a set of $N$ pre-NMS or raw bounding boxes $\mathbf{B} = \{\text{bbox}_{n} = ({\mathcal{B}}_{n},{\mathcal{P}}_{n})\}_{n=1}^N$, \big(${\mathcal{B}}_{n} = \{\mathbf{o}_n,\mathbf{s}_n\}$, ${\mathcal{P}}_{n} = \{\ell_n, p_n\}$\big), where $\mathbf{o}_n = (x,y)$ is the center of the $n$-th bounding box, $\mathbf{s}_n$ indicates its shape information (including height $h$ and width $w$, and optional rotation orientation $\theta$ for rotated objects in DOTA), $\ell_n$ is classification label and $p_n \in [0,1]$ is classification score including background. In Hiding Attack\cite{jia2022fooling,yin2022adc}, an adversarial example $\mathbf{x}^{adv}$ should be as similar as original input $\mathbf{x}$ while the outputs $\mathbf{B}^{adv} = \{\text{bbox}_n^{adv}\}_{n=1}^N$ are far away from both original predictions $\mathbf{B}^{org}$ and  ground truth\footnote{We use ground truth (GT) like previous works for fair comparisons, but you can replace that with clean predicts (CP) for better results in this paper. \EG, when replacing GT with CP, mAP$_{50}$ drops from 5.9 to 2.1 against TOOD and drops from 5.2 to 4.3 against S$^{2}$A-Net} $\mathbf{B}^{gt}$ in both aspects of geometric information and classification labels. Previous works\cite{DAG,RAP,CA,CWA,TOG} formulate attacks as single optimization problems. They only optimize the attack loss and clip corresponding gradients to a small budget. 

Differently, the problem of adversarial attack is formulated as a joint optimization problem by minimizing attacking loss and the difference between clean inputs $\mathbf{x}$ and adversarial counterparts in this paper $\mathbf{x}^{adv} = \mathbf{x} + \gamma^*$.
\begin{equation}
\hspace{-0.2cm}
    \gamma^* = \min_{\gamma}\{\lambda_{1} \mathcal{L}(\mathbf{B}^{org},\mathbf{B}^{gt},\mathbf{B}^{adv}) + \lambda_{2} \mathcal{D}(\mathbf{x}, \mathbf{x}+\gamma)\} 
\label{eq:overall}
\end{equation}
where $\mathcal{D}(\mathbf{x}, \mathbf{x}+\gamma)$ measures the perceptibility distance between two arguments. Specific optimization loss $\cal{L}$ should be considered to achieve the goal of attacking. $\lambda_{1}$ and $\lambda_{2}$ are used to weigh attack strength and the imperceptibility of perturbations.

\section{Methodology}
Three ingredients are essential against ODs: (\emph{i}) an structure that generates targets to be attacked; (\emph{ii}) an attacking strength loss that pushes clean predicts to adversarial objective; (\emph{iii}) a constraint loss that controls the magnitude and distribution of adversarial perturbations. The overall structure of LGP  is illustrated in \cref{fig:pic/overall.pdf} and \cref{alg: algorithm1}. 

\begin{figure}[!t]
\centering
\includegraphics[width=.9\linewidth]{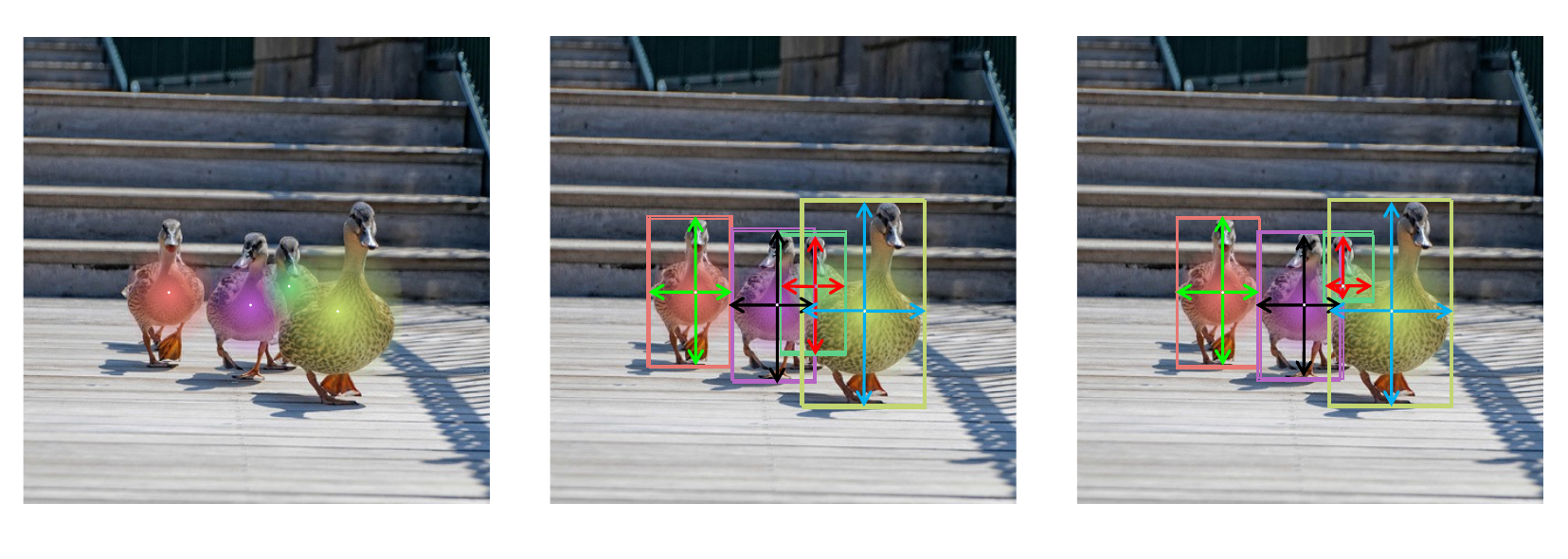}
  \caption{\textbf{Assigner.} 
  \label{fig: Heatmap and Assigner} The left is an object-wise Heatmap for imperceptibility. The middle/last shows attacked target boxes generated by Assigner based on IoU / scores.}
\end{figure}

\smallskip
\subsection{Assigner Towards To Generic Attacks}\label{sec:target generating}

For high-quality adversarial targets, most previous attacks\cite{DAG,RAP,CA,CWA,wang2021daedalus} tend to leverage some particular components of detectors, such as anchor\cite{faster-rcnn, TOG, CA, CWA}, RPN\cite{faster-rcnn, DAG, RAP}, RoI heatmap\cite{faster-rcnn, TOG}, or NMS\cite{NMS,wang2021daedalus}. This hinders their generalization from launching an attack for new detectors because different detectors have different components.
For a unified attack, we decouple the attack from the intrinsic structure of ODs and only consider their outputs before post-processing. But thousands of outputs (especially in one-stage detectors) bring an unbearable computational overhead, exposing a new question: \textbf{which part of outputs should be attacked?} In other words, we need to select adequately meaningful targets $\cal{T}$ to attack\cite{DAG}. To this end, we present a \emph{Trackable Target Assignment} strategy, which selects high-quality original targets ${\cal{T}}_{org}$ from pre-NMS proposals $\mathbf{B}^{pre}$ and tracks actively the best match sets with them as targets awaiting attack ${\cal{T}}$. 

\textbf{Assigner.} Plentiful random proposals bring uncertain attention, which inevitably focuses on a part of objects with many proposals while omitting the remained objects (\ie, different objects have different amounts of proposals in the same iteration, termed as \textbf{uncertainty}, \eg, the left three pictures in \cref{Fig:pic/1/Assigner.pdf}). To solve this problem, we assign averagely high-quality proposals to each ground truth (\eg, the last picture in \cref{Fig:pic/1/Assigner.pdf}), resulting in balanced original targets that serve as a foundation for later generations of adversarial targets. Specifically, LGP first assigns a fixed number of top $N_{i}$ proposals which have high IoU rates with ground truth. In this case, most high-quality proposals are considered (\eg, the middle of ~\cref{fig: Heatmap and Assigner}), but some proposals with lower IoU and high confidence should also be considered (\eg, a duck whose body is overlapped in \cref{fig: Heatmap and Assigner}). Therefore, we introduce the second criterion that further assigns some top $N_s$ proposals sorted by classification scores (\eg, the right of ~\cref{fig: Heatmap and Assigner}). After that, each ground truth has corresponding 
 one-to-many original proposals.

\textbf{Tracking.} Because the neighbor pixels of previous correct boxes are changed in different iterations, another sub-optimal bounding box may be detected around the attacked one \cite{larget-scale} (\ie, the same object has different proposals in a different iteration, termed as \textbf{instability}). To solve this instability, we construct attacked targets ${\cal{T}}_{i}$ according to the similarity between fixed ${\cal{T}}_{org}$  and changeable $\mathbf{B}^{pre}_{i}$ in the $i$-th iteration. Specifically, we use the best match proposals in $\mathbf{B}^{pre}_{i}$ which have top IoU rates and scores with ${\cal{T}}_{org}$ as ${\cal{T}}_{i}$. By now, LGP forces an one-to-one matching between ${\cal{T}}_{org}$ and ${\cal{T}}_{i}$, while maintaining an one-to-many mapping between ground truth and stable ${\cal{T}}_{i}$ boxes in different iterations. These mappings allow us to optimize the entire attack from an object-wise standpoint. 

\begin{algorithm}[!t]
	\caption{\label{alg: algorithm1} Local-Global Perturbations(LGP)}
	\KwIn{original image $\mathbf{x}$;\\
  \quad \qquad the detector ${\mathcal{D}et}(x) \xrightarrow[]{pre-\textrm{NMS}} \mathbf{B}^{pre}$\\
  \quad \qquad the ground truth ${\mathbf{B}}^{gt}$\\
  \quad \qquad the maximal iterations $\emph{I}_0$\\
  }
	\KwOut{Adversarial Examples $\mathbf{x + \gamma}^{*}$}
    Initialize:\, ${\mathcal{T}}_{org}  \leftarrow Assigner({\mathbf{B}}^{pre}, {\mathbf{B}}^{gt})$\,\\
    \While{$i \leq I_0$ and ${\mathbf{B}}^{gt}_{i} \neq \varnothing$}{
	    ${\cal{T}}_{i}  \leftarrow Tracking\{{\mathcal{D}et}(\mathbf{x} + \gamma_{i-1}), {\cal{T}}_{org}\}$\\
	    ${\mathcal{T}}_{i}^{\prime}  \leftarrow Attacker({\mathcal{T}_{i}},{\mathbf{B}^{gt}})$ \\
	    $({\cal{H}}_{i}, {\mathbf{B}}^{gt}_{i})  \leftarrow Adaptor({\mathcal{T}_{i}},{\mathbf{B}^{gt}})$\\
	    $Loss_{i}  \leftarrow {\cal{L}}({\cal{T}}_{i},{\mathcal{T}}_{i}^{\prime}) + {\cal{D}}(\mathbf{x},\mathbf{x}+\gamma_{i-1}) \cdot {\cal{H}}_i$ \\
	    ${x}^{adv}_{i} \leftarrow Optimizer(\mathbf{x}+\gamma_{i-1},Loss_{i})$ \\
	    $i \leftarrow{i+1}$ 
		}
\end{algorithm}

\subsection{Attacker Guided by High-Level Objective\label{sec:Attacker}}

Multi-task attacks are more powerful\cite{RAP} and help more security scenes\cite{zhang2019towards,larget-scale} than single-task attacks. Thus, we then try to strengthen our attack by leveraging the multi-task nature of ODs to attack classification and regression simultaneously. However, the gradients of multi-task attacks are not fully aligned\cite{zhang2019towards}, impeding subsequent optimization. For an aligned multi-task attack, we set a unified objective ``Hiding Attack (HA)"\cite{jia2022fooling} to guide our design of losses for blinding ODs.
 Specifically, we argue that a good adversarial example should be able to minimize the difference between predicts and background from the perspective of shapes, locations, and semantics of proposals. More important, high-level semantics (HA) are more meaningful than previous untargeted attacks.

Denoted ${\mathcal{T}}_{i} = \{\text{b}_{n} =(\mathbf{o}_n,\mathbf{s}_n,\ell_n, p_n)\}_{n=1}^N$ is the selected targets for attacking in $i$-th iteration. 

\emph{Shape Constraint}: Motivated by common sense ``big objects have big boxes", we try to make big objects smaller and vice versa. In other words, we hope to provide incorrect geometry information by adding a scaling ratio $\zeta$ to expand or shrink bounding boxes and then hide true-positive proposals from the eyes of ODs. Detaily, we use Smooth L1 (SL1) to decrease 
 the difference between selected targets $b_{n}$ and adversarial targets $b_{n}^{\prime} = [\mathbf{o}_n,\mathbf{s}_n^{\prime} = (w_n^{\prime},h_n^{\prime},\theta_n), \ell_{n},p_{n}]$ ($w_{n}^{\prime} = \zeta w_{n},h_{n}^{\prime} = \zeta h_{n}$). Afterward, we can push the shape distribution of original targets to configured adversarial targets by our shape loss ${\cal{L}}_{shape}$.
\begin{equation*}
\begin{aligned}
  d = SL1(m,n)=
\left\{  
             \begin{array}{cc}  
              \frac{1}{2}(m - n)^{2}, \lvert m - n \rvert< 1.0   \\   
             \lvert{m - n}\rvert - \frac{1}{2}, otherwise  
             \end{array}  
\right.
\end{aligned}
\end{equation*}
\begin{equation}
\begin{aligned}
  {\cal{L}}_{shape}(b_n,b_{n}^{\prime}) =  d(w_n,w_{n}^\prime) + d(h_n,h_{n}^\prime)
\end{aligned}
\label{eq:shape attack}
\end{equation}
\emph{Localization Constraint}: As for the goal of hiding the location of objects, generated AEs should lead to non-meaningful localization outputs. That is to say, its outputs should be far away from any foreground pixels. We keep predictions far from ground truth by the IoU distance and center-point offsets as the location loss ${\cal{L}}_{loc}$.
\begin{equation}
\begin{aligned}
  {\cal{L}}_{loc}(b_n,b_n^{gt}) = \text{IoU}(b_n,b_n^{gt})  - d(\mathbf{o}_n,\mathbf{o}_n^{gt})
 \label{eq:location attack}
\end{aligned}
\end{equation}
where $b_n^{gt} = (\mathbf{o}_n^{gt},\mathbf{s}_n^{gt})$ is assigned by many-to-one mapping in \cref{sec:target generating}. Every predicts $b_n$ have a unique $b_n^{gt}$.

\emph{Semantic Constraint}: In order to hide the semantic information in AEs, we expect the output classification labels $\ell_n^{adv} = \varnothing$, where $\varnothing$ indicates the background or ``no object" label. Thus, we minimize semantic loss ${\cal{L}}_{cls}$ by Logit Loss\cite{logit} or Cross-Entropy Loss (CE). CE will be used if there is no background probability in some detectors.

\begin{equation}
\begin{aligned}
  {\cal{L}}_{CE}(p_n,\ell_{b}) = -log(\frac{e^{z_b}}{\sum e^{z_j}}) =-z_b + log(\sum{e^{z_j}})
\end{aligned}
\label{eq: CE attack}
\end{equation}
\begin{equation}
\begin{aligned}
  {\cal{L}}_{Logit}(p_n,\ell_{b}) = -z_{b}
  \label{eq:3.4.2}
\end{aligned}
\end{equation}
where $z_b$ is the probability of the background label and $z_j$ is the probability of different labels.

Finally, each target $t_n$ is assigned a bigger or smaller bounding box $b_n^{\prime}$ for shape attack, a ground truth $b_n^{gt}$ for location attack, and a background class label $\ell_{b}$ for classification attack. Thus, we can construct the adversarial targets $t_n^{\prime}= \{b_n^{\prime},b_n^{gt},\ell_{b}\}\in{\mathcal{T}^{\prime}_{i}}$ and further specify the attacking loss function ${\cal{L}}$ as below:
{\setlength\abovedisplayskip{3pt}
\setlength\belowdisplayskip{3pt}
\begin{equation}
\begin{aligned}
    {\cal{L}} ~ = ~ &\sum_{n=1}^{N} ~ {\alpha  
  {\cal{L}}_{shape}(b_n,b_n^{\prime})} ~ / 
  ~ N ~ + ~\\& \sum_{n=1}^{N}~ \beta  {\cal{L}}_{loc}(b_n,b_n^{gt}) ~ / ~ N ~+~ \sum_{n=1}^{N} ~ \tau  {\cal{L}}_{cls}(p_n,\ell_{b})
  \label{eq:Attacker}  
\end{aligned}
\end{equation}}

\subsection{Limiter Guided by Object-Wise Controllability\label{sec:Limiter}}

\textbf{What adversarial perturbations should be generated in object detection?} Influenced by image-level clipping of perturbations against classifiers, existing works against detectors also generate final AEs from the perspective of the global image. Since ``objects" are key in object detection, non-informative areas (\eg, backgrounds) should be excluded, encouraging perturbations to attach to meaningful pixels and decreasing the computational overhead in other regions. In short, \textbf{Not all pixels are what you need}. Although attacking stable high-quality proposals can be seen as a rough local attack, there is still a risk to get uncontrollable perturbations without supervision\cite{frequency}. Note that controllability is not imperceptibility, the former denotes the magnitude, position, and distribution of perturbations are controllable by the attacks, but the latter denotes the magnitude of perturbations is small.


 \begin{figure}[!t]
  \centering
\includegraphics[width=0.98\linewidth]{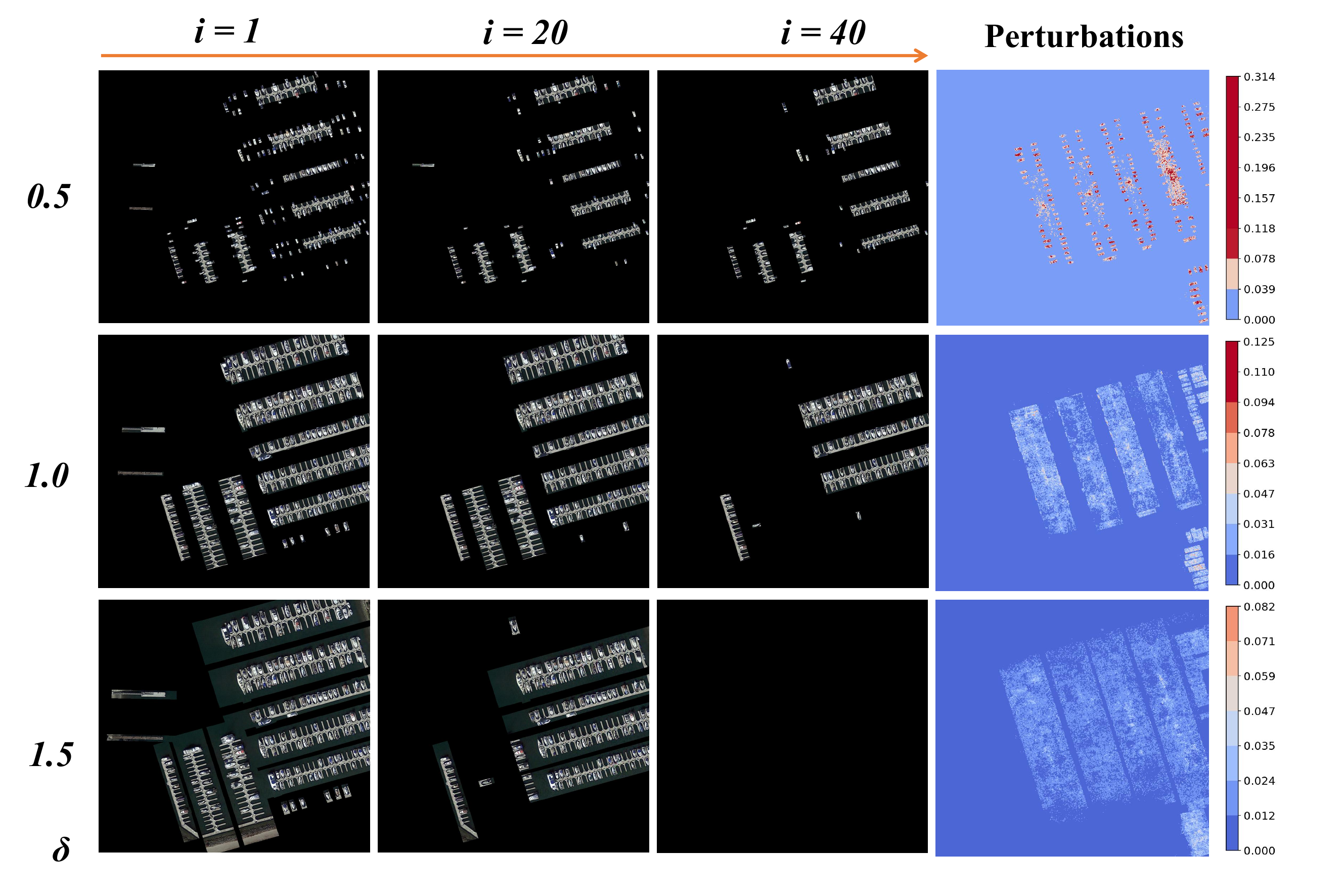}

  \caption{ In different iterations, the Adaptor limits perturbed regions in an object-wise constraint (\ie, the first three columns). The last column shows the final perturbations. As shown, the more space is disturbed, the more objects are successfully attacked in the same iteration. \label{fig:fore}}
  \vspace{0.1cm}
\end{figure}
\textbf{Foreground-Background Separation (FBS).\label{sec:FBS}} Deep networks focus implicitly on the objects by their attention mechanism, which motivates us to control the distribution of perturbations from the object wise. But a simple mask for excluding backgrounds could destroy learned perturbations leading to a weak attack (\eg, when PGD$_{reg}$ attacks Oriented R-CNN with a mask to foregrounds, mAP$_{50}$ increases from 10.0 to 54.4, but the one of LGP from 4.0 to 19.0). Thus, we design a novel constraint for the joint optimization with the Attacker, limiting perturbations according to the location and shape of objects. In this way, we construct an object-wise Heatmap {${\mathcal{H}}(\mathbf{B}^{gt})$} as a priori limit to control the random distribution of perturbations. We use a simple but efficient Euclidean distance for every bounding box in this paper (\eg, the left picture in \cref{fig: Heatmap and Assigner}).

\begin{equation}
\label{eq:object-wise Heatmap}
 {\cal{H}}({\mathbf{B}}^{gt}) = \eta  \left\{
\begin{aligned}
&\sqrt{\frac{(x-x_c)^{2}+(y-y_c)^{2}}{w^{2} + h^2}}, \, (x,y) \in \delta ~ {\mathbf{B}}^{gt}   \\   
             &\quad \quad \quad  1.0,\quad otherwise 
\end{aligned}
\right.
\end{equation}
where ${\cal{H}}$ denotes the object-wise Heatmap, (x,y) is the coordinates of any points, ($x_{c}$,$y_{c}$) is the center coordinates of ground truth, $\delta~ {\mathbf{B}}^{gt}$ is the perturbed space with a scaling ratio $\delta$. When (x,y) is located in any $\delta ~{\mathbf{B}}^{gt}$, we give it gaussian weights to limit allowed areas for perturbations.

\textbf{Adaptor.\label{sec:Adaptor}} To avoid a suboptimal result, we update limited regions adaptively to improve the flexibility of the Limiter. In this way, we can further guide perturbations for an \textbf{object-wise} representation. Given the ground truth ${\mathbf{B}}^{gt}$, we think the successful-attack objects ${\mathbf{B}}^{gt}_{success}$ mean corresponding predictions of them are background (according to the sorted IoUs and scores), and others are failed-attack objects ${\mathbf{B}}^{gt}_{i}$. In this paper, we cancel the constraints in the regions of ${\mathbf{B}}^{gt}_{success}$. \cref{fig:fore} shows the change of limited failed-attack foregrounds with different scale $\delta$ in different iterations. 
We can clearly see that \textbf{Adaptor tells adaptively the Limiter where to limit.}

Based on the above settings, we can formulate the distance metric $\cal{D}$ of LGP in $i$-th iteration as below:
\begin{equation}
\begin{aligned}
  {\cal{D}}_{i} = &{\cal{D}}(\mathbf{x},\mathbf{x + \gamma}_{i-1})_{i} \cdot {\cal{H}}_{i} = d(\mathbf{x}^{bg},(\mathbf{x + \gamma}_{i-1})^{bg})+\\&d(\mathbf{x}_{i}^{fg} \cdot {\cal{H}}_{i},({\mathbf{x + \gamma}_{i-1}})_{i}^{fg} \cdot {\cal{H}}_{i})+ \epsilon~ \ell_{2}({\gamma}_{i-1} \cdot {\cal{H}}_{i})
\end{aligned}
\label{eq:dynamic constranit}
\end{equation}
where $\mathbf{x}^{bg}$ denotes the background of clean images, $\mathbf{x}_{i}^{fg}$  denotes the failed-attack foregrounds of clean images in the $i$-th iteration (likewise for $(\mathbf{x + \gamma}_{i-1})^{bg}$, $(\mathbf{x}+ \gamma_{i-1})^{fg}_{i}$ for AEs). For convenience, ${\cal{H}}_{i}$ denotes ${\cal{H}}({\mathbf{B}}^{gt}_{i})$.  We apply $\ell_{2}$ norm to limit perturbations $\gamma$, which is widely used in the attack of image classification\cite{CW}. $d(\cdot)$ is in \cref{eq:shape attack}.

\section{Experiments}
\subsection{Experimental Setup}
{\bf Datasets.} We evaluate the performance of our method on two popular datasets: MS-COCO\cite{COCO} for horizontal bounding boxes and DOTA-v1.0\cite{DOTA} for rotated bounding boxes. MS-COCO is challenging to split due to the overlap among objects and DOTA-v1.0 has many small crowded instances indicating a low tolerance for perturbations. We attack their validation sets for a fair comparison.

{\bf Victim Detectors.} We select respectively eight representative detectors as victim models on two datasets. On DOTA-v1.0, we attack Oriented R-CNN (OR)\cite{OrientedRCNN}, Gliding Vertex (GV)\cite{xu2020gliding}, RoI Transformer (RT)\cite{RoITransformer}, ReDet (RD)\cite{ReDet} as two-stage detectors, Rotated Retinanet (RR)\cite{RetinaNet}, Rotated FCOS (RF)\cite{FCOS}, S$^{2}$A-Net\cite{S2ANet} as single-stage detectors, and AO2-DETR (AD)\cite{AO2DETR} based on transformer\cite{transformer}. On MS-COCO, we attack Faster R-CNN (FR)\cite{faster-rcnn}, Cascade R-CNN (CR)\cite{CascadeRCNN}, Sparse R-CNN (SR)\cite{SparseRCNN} and SABL Faster R-CNN (SABL)\cite{sabl} as two-stage detectors, RepPonits(RP) \cite{reppoints}, VFNet\cite{vfnet}, TOOD\cite{tood} as single-stage detectors, and Deformable DETR (D.DETR)\cite{DeformableDetr} based on transformer\cite{transformer}. ResNet\cite{resnet} or ResNeXt\cite{resnext} are their backbones (R50, R101, and X101 are ResNet50, ResNet101, and ResNeXt101). The above models and codes are implemented based on the open-source mmdetection\cite{mmdetection} and mmrotate\cite{mmrotate} library.

{\bf Evaluation metrics.} We use mean average accuracy (mAP) with IoU threshold 0.5 and the initial number of attacked targets $\textbf{N}_{T}$ per image to evaluate the attack ability for a fair comparison. Besides, we introduce the total number of predicted boxes with IoU threshold 0.75 $\textbf{N}_{75}$ to evaluate the success rate of the Hiding Attack.  To reflect the imperceptibility, we choose three different metrics in PIQ\cite{piq}, including IW-SSIM\cite{IWSSIM}, PSNR-B\cite{PSNRB}, and 
FID\cite{FID} to evaluate the distance between clean and perturbed images. We multiply the value of IW-SSIM and mAP$_{50}$ by 100 for a clear comparison. We evaluate the time-consuming of all attacks on the TITAN X (PASCAL) machine.

{\bf Parameters Setting.}  We use Adamax\cite{Adam} with a learning rate of 0.1 for 50 iterations per image.
$\lambda_{1}$, $\lambda_2$ are respectively 1.0,  0.1 for attack strength and imperceptibility in \cref{eq:overall}. The Assigner assigns five bounding boxes (${\mathbf{N}}_{i}$) based on IoU and five bounding boxes (${\mathbf{N}}_{s}$) based on scores. The default values for $\alpha,\beta,\tau$ in \cref{eq:Attacker} are 1.0, and $\zeta$ in \cref{eq:shape attack} is 3 in DOTA and 0.1 in MS-COCO. $\delta$ in \cref{eq:object-wise Heatmap} is 1.5 and $\epsilon$ in \cref{eq:dynamic constranit} is 0.1. \textbf{All attacked detectors use the same parameters without crafted adjustments.}

\begin{table*}[!htbp]
\caption{\textbf{The comparisons use different adversarial attack methods.} PGD$_{cls}$ and PGD$_{reg}$ denote that attacking the pre-NMS $\mathbf{B}^{pre}$ by classification scores and location offsets for RPN-based ODs. Others are modified slightly to fit different detectors and datasets for a better result (* is extracted from \cite{RAP}). $^{\diamond}$ denotes we modify CWA\cite{CWA} for launching an attack for pre-NMS outputs of two-stage detectors with its class-wise loss. TOG is TOG with vanishing loss in\cite{TOG}.  $^{\dagger}$ denotes the results of LGP with 10 iterations. $^{\ddagger}$ denotes the results of LGP with 150 iterations.Time is the average time to generate an adversarial example. N$_{T}$ is the number of initial targets to be attacked. We highlight the best and second best results by \textcolor{red}{red} and \textcolor{blue}{blue}. As shown, LGP has the best attacking capacity and imperceptibility while using the least proposals.\label{table:baseline}}
\resizebox{\textwidth}{!}{
\begin{tabular}{ccccccccccccccc}
\hline \hline
\multirow{2}{*}{Methods} & \multirow{2}{*}{Budgets$\downarrow$}  &\multicolumn{6}{c}{Faster R-CNN\cite{faster-rcnn}}                               &  & \multicolumn{6}{c}{Oriented R-CNN\cite{OrientedRCNN}} \\ \cline{3-8} \cline{10-15} 
        &         & IW-SSIM$\downarrow$          & PSNR-B$\uparrow$           & FID$\downarrow$            & mAP$_{50}$$\downarrow$  &$\textbf{N}_{T}$$\downarrow$ &Time(s)$\downarrow$ & & IW-SSIM$\downarrow$  & PSNR-B$\uparrow$  & FID$\downarrow$   & mAP$_{50}$ $\downarrow$&$\textbf{N}_{T}$$\downarrow$ &Time(s)$\downarrow$ \\ \hline
CLean    &         &            &             &             & 51.0 &  &  &    &    &   & & 83.3& &    \\
PGD$_{cls}$\cite{PGD}     &8        & 1.15           & 35.6             & 3.53            & 3.4 & 2000 &6.21 & &1.66   &36.1    & 3.27  & 14.3&2000 &13.3\\
PGD$_{reg}$\cite{PGD}      &8       & 1.54           & 34.4             & 5.12            & 2.4 & 2000& 5.37   & &1.92   & 35.2    & 4.26  & 10.0 &2000  & 13.78  \\
DAG\cite{DAG}     &8     & 0.95           & 40.0             & 4.56            & 3.3 &115&  13.7  &   & 0.576    & 45.2    & 1.04  & \textcolor{blue}{4.5} &555  & 24.2 \\
RAP\cite{RAP}$^{*}$     &$\backslash$      & $\backslash$ & $\backslash$ &$\backslash$  & $10.5^{*}$ &2000&  $\backslash$  & & 1.13   & 39.4    & 1.44  & 9.5 &2000   & 39.0  \\
CWA\cite{CWA}$^{\diamond}$     & 8       & 1.64          & 35.3             &8.49          &7.7 &\textcolor{blue}{60} & 9.2   &  & 1.53  & 39.4    &1.53  &9.4 &\textcolor{blue}{485}  & 8.4 \\
TOG\cite{TOG}     &16       &0.40          & 39.2             &\textcolor{blue}{1.70}          &3.3 &2000 & \textcolor{blue}{3.2}   &  &0.49  &40.5    &0.645 &12.9 &2000  & 8.55 \\
\textbf{LGP(ours)}   &optimize        & \textcolor{blue}{0.52}          & \textcolor{blue}{40.7}             & 1.96          &    \textcolor{red}{1.5} & \textcolor{red}{34}& 3.61 &    & \textcolor{blue}{0.222}  & \textcolor{blue}{47.3}    & \textcolor{blue}{0.268}  & \textcolor{red}{4.0} &\textcolor{red}{112} & \textcolor{blue}{6.12} \\ 
\textbf{LGP$^{\dagger}$(ours)}   &optimize        & 1.75          &36.5             & 5.02          &\textcolor{blue}{2.3} & \textcolor{red}{34}& \textcolor{red}{1.96} &    & 0.705  & 42.63    & 2.74   & 7.6 &\textcolor{red}{112} & \textcolor{red}{5.3} \\
\textbf{LGP$^{\ddagger}$(ours)}   &optimize        & \textcolor{red}{0.179}          & \textcolor{red}{43.6}             & \textcolor{red}{1.00}          &\textcolor{red}{1.5} & \textcolor{red}{34}& 8.59 &    & \textcolor{red}{0.064}  & \textcolor{red}{50.6}    & \textcolor{red}{0.101}  & 5.2 &\textcolor{red}{112} & 25.5 \\ \hline

\multirow{2}{*}{Methods} & \multirow{2}{*}{Budgets$\downarrow$}  &\multicolumn{6}{c}{RepPoints\cite{reppoints}}                               &  & \multicolumn{6}{c}{S$^2$A-Net\cite{S2ANet}} \\ \cline{3-8} \cline{10-15} 
        &         & IW-SSIM$\downarrow$          & PSNR-B$\uparrow$           & FID$\downarrow$            & mAP$_{50}$$\downarrow$  &$\textbf{N}_{T}$$\downarrow$ &Time(s)$\downarrow$ & & IW-SSIM$\downarrow$  & PSNR-B$\uparrow$  & FID$\downarrow$   & mAP$_{50}$ $\downarrow$&$\textbf{N}_{T}$$\downarrow$ &Time(s)$\downarrow$ \\ \hline
CLean    &         &            &             &             &51.8 &  &  &    &    &   & & 81.2& &    \\
PGD$_{cls}$\cite{PGD}     & -        & -           & -             & -            &- & - &- & & -   & -    & -  &-&- & -\\
PGD$_{reg}$\cite{PGD}       & -        & -           & -             & -            &- & - &- & & -   & -    & -  &-&- & - \\
DAG\cite{DAG}      & -        & -           & -             & -            &- & - &- & & -   & -    & -  &-&- & -\\
RAP\cite{RAP}      & -        & -           & -             & -            &- & - &- & & -   & -    & -  &-&- & - \\
CWA\cite{CWA}     & 8        & 1.22           & 40.5            &4.29           &\textcolor{red}{2.4} &4008 & 8.33   &  & 1.68   & 35.8    & 2.94  & 11.8 &\textcolor{blue}{3313}   & 14.26 \\
TOG\cite{CWA}     & 16        & \textcolor{red}{0.39}   & 39.2    & \textcolor{red}{1.5}  &10.6 &\textcolor{blue}{1002}   & \textcolor{blue}{2.93}   &  & 0.504   & 40.3    & 0.657  &20.2 &5344   & \textcolor{red}{5.21} \\
\textbf{LGP(ours)}   &optimize        & 0.67          & \textcolor{blue}{41.1}             & \textbf{2.3}          &  5.0 & \textcolor{red}{100}& 21.3 &    & \textcolor{blue}{0.239}  & \textcolor{blue}{47.0}    & \textcolor{blue}{0.317}  & \textcolor{blue}{5.2} &\textcolor{red}{287} & 28.1 \\
\textbf{LGP$^{\dagger}$(ours)}   &optimize        & 1.49          & 38.6             & 3.83          & 13.6 & \textcolor{red}{100}& \textcolor{red}{2.77} &    & 0.809  & 42.9    & 0.774  & 10.8 &\textcolor{red}{287} & \textcolor{blue}{9.38} \\
\textbf{LGP$^{\ddagger}$(ours)}   &optimize        & \textcolor{blue}{0.53}          &\textcolor{red}{41.65}             & \textcolor{blue}{2.17}          &  \textcolor{blue}{3.0} & \textcolor{red}{100}& 47.26 &    & \textcolor{red}{0.163}  & \textcolor{red}{47.6}    &\textcolor{red}{0.229}  & \textcolor{red}{4.2} &\textcolor{red}{287} & 101.8 \\
\hline
\end{tabular}}
\end{table*}

\begin{table*}[!htbp]
  \centering
\caption{\textbf{LGP attacks different detectors  on MS-COCO (left) and DOTA-v1.0 (right)}. ``clean" and ``adv" are respectively results before the attack and after the attack. $\textbf{N}_{75}$ denotes the number of predicts with IoU threshold 0.75. In this table, all attacks use \emph{the same hyperparameters} which indicates the generic attack capacity of LGP.
\label{table:LGP in different detectors.}}
  \resizebox{\textwidth}{!}{
\begin{tabular}{cccccccc|cccccccc}
\hline \hline
\multirow{2}{*}{MS-COCO}      & \multirow{2}{*}{Backbone}    & \multirow{2}{*}{FID $\downarrow$}  & \multicolumn{2}{c}{mAP$_{50}$} & & \multicolumn{2}{c|}{$\textbf{N}_{75}$}   & \multirow{2}{*}{DOTA}           & \multirow{2}{*}{Backbone}                    & \multirow{2}{*}{FID $\downarrow$}                                       & \multicolumn{2}{c}{mAP$_{50}$}  & & \multicolumn{2}{c}{$\textbf{N}_{75}$}              \\   \cline{4-5} \cline{7-8}  \cline{12-13} \cline{15-16} 
                              &            &           & clean $\uparrow$          & adv $\downarrow$ & & clean                 & adv $\downarrow$          &  &                                                                              &  & clean $\uparrow$                 & adv $\downarrow$ & & clean                & adv $\downarrow$                 \\
\hline \multirow{3}{*}{FR\cite{faster-rcnn}} & R50   & 1.96       & 51.0           & 1.5          &  &23053 &2496 & \multirow{3}{*}{OR\cite{OrientedRCNN}} & \multirow{3}{*}{R50}   & \multirow{3}{*}{0.268}   & \multirow{3}{*}{83.3} & \multirow{3}{*}{4.0} && \multirow{3}{*}{39341}&\multirow{3}{*}{4055}\\
                              & R101           & 2.43       & 53.0           & 1.6          &  &23089  & 2457                                &                           &  &                          &                       &                       &  &                        &                      \\
                              & X101          & 2.50       & 55.2          & 1.5          &  & 22856 & 2249                                &                                                   &                       & &&                      &  &                        &                        \\
CR\cite{CascadeRCNN}                 & R50        & 2.33       &51.3          & 0.8          &  &23274 &1821  & GV\cite{xu2020gliding}                & R50                                 & 0.206                    & 81.3                  & 23.1 &&26974&10092                 \\
SABL\cite{sabl}            & R50        & 1.589       & 50.7          & 3.1          &  &26949 &6482 & RT\cite{RoITransformer}                & R50                         & 0.221                    & 86.5                & 20.8  &&37730&    11703            \\
SR\cite{SparseRCNN}                  & R50          & 1.80       & 47.6          & 10.4          &  &81370 &15270 & RD\cite{ReDet}                           & R50                                & 0.173                   & 83.3                  & 22.0   &&38430&   12564            \\
RP\cite{reppoints}                     & R50           & 2.30       & 49.0           & 5.0          &  &49830 &2468 & RR\cite{RetinaNet}               & R50                         & 0.429                  & 74.7                  & 10.2  &&100123&9723                \\
TOOD\cite{tood}                          & R50          & 3.32       & 51.8          & 5.9          &  &49780 &3783 & RF\cite{FCOS}                    & R50                           & 0.892                    & 78.7                  & 4.9   &&73288&5689               \\
VFNet\cite{vfnet}                         & R50           & 1.53       & 51.3          & 11.2          &  &56222 & 8929& S$^{2}$A-Net\cite{S2ANet}                          & R50                               & 0.317                  & 81.2                  & 5.2   &&68338&6259               \\
D.DETR\cite{DeformableDetr}               & R50           & 1.36       & 60.7          & 12.3          &  &70048 &24513 & AD\cite{AO2DETR}                        & R50                     &  0.416                   & 85.0                  & 8.6     &&198477&52250             \\ \hline \hline
\end{tabular}}
\end{table*}

\subsection{White-box Attacks}

In this section, we quantify the \textbf{effectiveness} of adversarial examples by mAP$_{50}$\&$\textbf{N}_{75}$ and their \textbf{imperceptibility} by three image quality assessments\cite{piq} on sixteen advanced detectors (attacking horizontal and rotated boxes). 

\begin{figure*}
\begin{minipage}{0.7\textwidth}
\tabcaption{\textbf{Ablation Study in Faster R-CNN (FR) \cite{faster-rcnn} / Oriented R-CNN (OR) \cite{OrientedRCNN}.}The three rows No.2-4 are different attacking loss with an image-level distance constraint. where $d_1 = d(\mathbf{x},\mathbf{x} + \gamma_{i-1})$, $d_2 = \ell_{2}(\gamma_{i-1})$, and $d$ in \cref{eq:shape attack}. We adjust the limited regions of Limiter for a better local attack in the next three rows. The last three rows are based on different original targets. \label{table:ablation study}}
\resizebox{\textwidth}{!}{
\begin{tabular}{cccccccccc}
\hline \hline 
 FR / OR         & ${\cal{L}}_{cls}$ & ${\cal{L}}_{shape}$& ${\cal{L}}_{loc}$                              & ${\cal{D}}(\cdot)_{i}$                                               &${\cal{T}}_{org}$ & FID$\downarrow$  & mAP$_{50}$ $\downarrow$  & $\textbf{N}_{75}$ $\downarrow$          \\ \hline 
1        & $\setminus$                      & $\setminus$                                     & $\setminus$       &$\setminus$ &$\setminus$ &$\setminus$&  51.0/83.3 &23053 / 39341     \\
2        &\checkmark &&                        & $d_{1}$                                      & HQ       &0.618 / 0.104  &  10.2 / 21.6 & 11338 / 17319      \\
3   &\checkmark & \checkmark &          & $d_{1}$                                        & HQ            &1.16 / 0.183  & 4.4 / 10.9 & 7376 / 13514         \\
4         &\checkmark & \checkmark & \checkmark  & $d_{1}$                                       & HQ           &1.20 / 0.184  & 2.4 / 10.5 & 3554 / 12738         \\
5  &\checkmark & \checkmark & \checkmark & $d_{1}$  + $\epsilon d_{2}$                           & HQ        &1.12 / 0.177  & 2.8 / 11.5  & 3806 / 13689        \\
6        &\checkmark & \checkmark & \checkmark  & ($d_{1}$ + $\epsilon d_{2}$) $\cdot {\cal{H}}$ & HQ             &1.57 / 0.195  & 1.8 / 5.7  & 2841 / 7903       \\ 
7       &\checkmark & \checkmark & \checkmark  &
\cref{eq:dynamic constranit}
& HQ      & 1.96 / 0.268  & \textbf{1.5} / \textbf{4.0} & \textbf{2496} / \textbf{4055}\\
8  &\checkmark & \checkmark & \checkmark  & 
\cref{eq:dynamic constranit}
& ${\mathbf{B}}^{pre}$        &2.18 / 0.338  & 3.1 / 5.7  & 4005 / 5462        \\
9 &\checkmark & \checkmark & \checkmark  &
\cref{eq:dynamic constranit}
& Predicts     &2.03 / 0.590  & 3.2 / 9.8 & 4086 / 8428\\
\hline \hline
\end{tabular}}
\end{minipage}
\hspace{0.2cm}
\begin{minipage}{0.25\textwidth}
  \includegraphics[width=\linewidth]{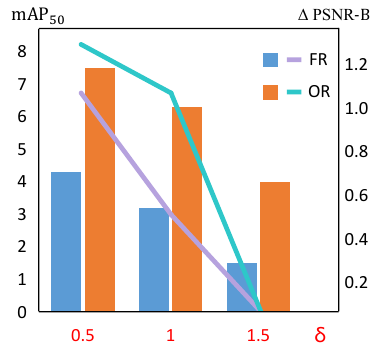}
\figcaption{\fontsize{7.5bp}{17bp} LGP with different fore scales $\delta$ in \cref{eq:object-wise Heatmap}. Histogram shows mAP$_{50}$ and the line chart shows PSNR-B minus its minimum value. \label{fig:fore_scale}}
\end{minipage}

\end{figure*}

\textbf{Comparisions.} \cref{table:baseline} reveals our attack is successful on two datasets. LGP outperforms most baselines\footnote{More comparisons shown in supplementary materials.} with the lowest mAP$_{50}$ and the best FID using the least initial targets. However, the main contributions of LGP are not crafting more powerful adversarial examples (AEs) with lower perceptibility, but generic (\ie, \cref{sec:target generating,sec:Attacker}) and controllable (\ie, \cref{sec:Adaptor}) attacks based on object-wise viewpoint. In \cref{table:baseline}, generic ability brings more time-consuming for one-stage detectors because the Assigner and Limiter need to select, assign, and split high-quality proposals from thousands of candidates in each iteration. But, LGP can launch a generic attack for new problems/datasets without changing strategies and hyperparameters with comparable strength and imperceptibility. In \cref{fig:fore,fig:fore_scale,fig:pic/1/per.pdf}, controllable ability shows that the smaller perturbed spaces are, the weaker strength of attacks will be. But, controllable perturbations are more meaningful for objects without the influence of the environment in the real world. 

Besides, we also provide the attack results with three different iterations 10, 50, and 150 in \cref{table:baseline}. We find that LGP gets comparable results with only 10 iterations. Due to the object-wise optimization, it reduces redundant perturbations as the number of iterations increases. Thus, the more iterations, the more high-quality adversarial examples.

\textbf{Generic attack:} Our first empirical observation is that \emph{different backbones have a negligible effect on high-intensity attacks and invisible perturbations. }LGP decreases the mAP$_{50}$ of Faster R-CNN by a large margin based on AEs trained with different backbones in \cref{table:LGP in different detectors.}. Meanwhile, their average FID is lower than 2.50, which means that generated perturbations are imperceptible.

Our second empirical observation is that \emph{LGP has generic attack capacity which is independent of ODs' structures and datasets}. In \cref{table:LGP in different detectors.}, LGP decreases about 90 percent mAP$_{50}$ with a better FID than other baselines in \cref{table:baseline}. We argue that generic capacity was given by attacking stable, high-quality proposals which decouple attack from the detectors' structure. Specifically, Assigner makes fixed original targets and tracks them for generating targets awaiting attack in abundant and changeable 
outputs\footnote{we visualize corresponding results in supplementary materials.} of ODs. Multi-task losses also improve the attack strength. 

\begin{figure*}[!t]
  \centering

    \includegraphics[width=.95\linewidth]{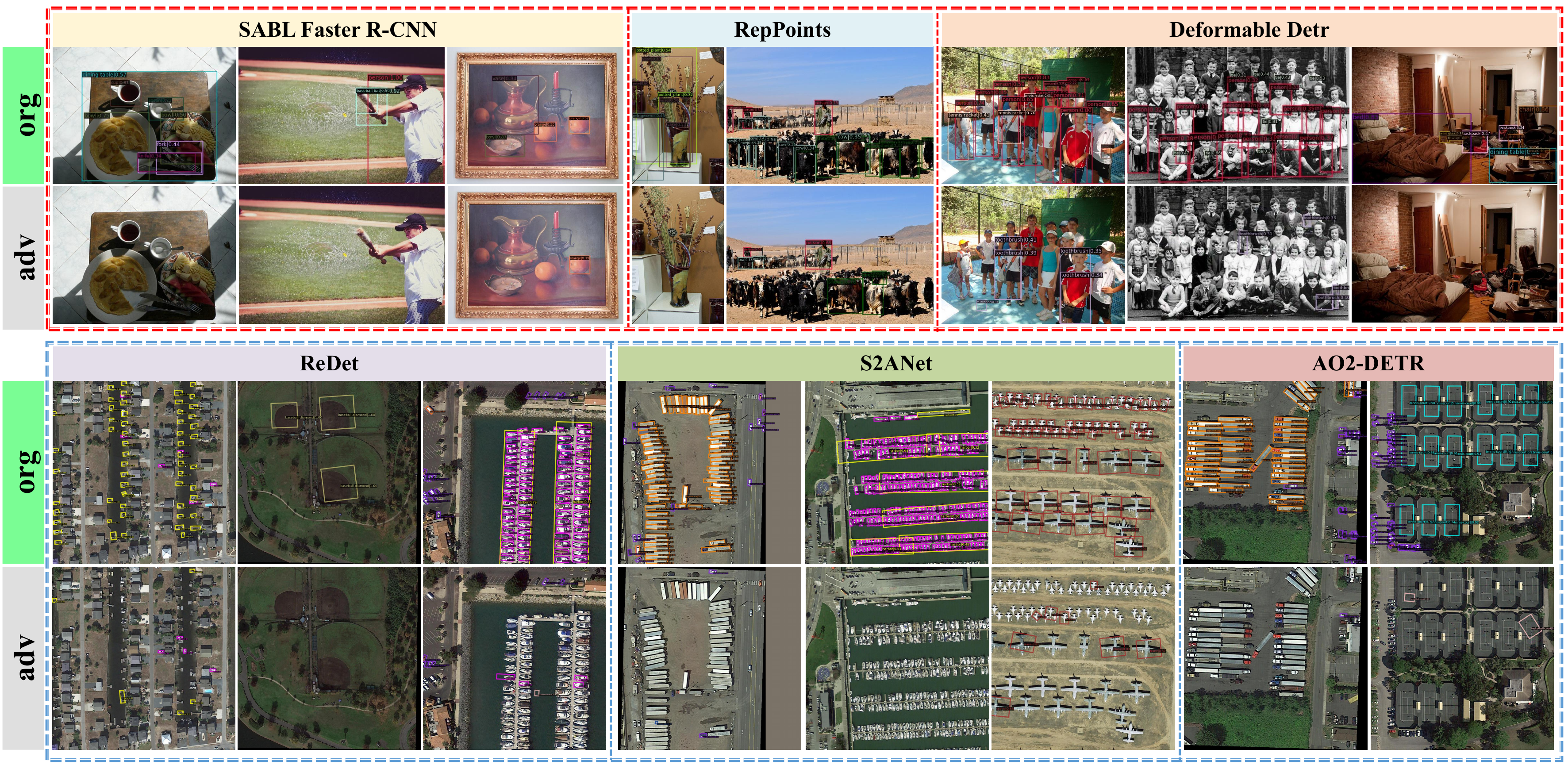}

  \label{fig:pic/More_results.pdf}
        \caption{More detected results by attacking two-stage detectors\cite{sabl,ReDet}, one-stage detectors\cite{reppoints,S2ANet}, and Transformer-based detectors\cite{DeformableDetr,AO2DETR} from left to right. The top \textcolor{red}{red box} is based on MS-COCO\cite{COCO}, and the bottom \textcolor{blue}{blue one} is based on DOTA\cite{DOTA}. As shown, LGP hides most objects without the influence of size, color, and density.}
\end{figure*}

The last empirical observation is that, \emph{high-level objective can decrease conflicts among heterogeneous losses.} Most predictions with high IoU values have been hidden successfully in \cref{table:LGP in different detectors.}.  This is thanks to a unified multi-objective optimization of Hiding Attack\cite{jia2022fooling}. In this case, we decrease the conflicts among different optimized branches and merge their attack space with a high-level semantic goal\footnote{Details in our ablation study and supplementary materials.}. We visualize more results in \cref{fig:pic/More_results.pdf}, which shows that our attack can hide different sizes and most types of objects.

\textbf{Controllable perturbations:} Different from clipping the perturbations to a small budget (this method only control the maximum of perturbations), we use proposal mappings and adaptive foreground-background splits to control the magnitude, position, and distribution of final perturbations. 

In \cref{Fig:pic/1/Assigner.pdf} and \cref{fig: Heatmap and Assigner}, LGP establishes a many-to-one relationship between targets awaiting attack and ground truth, ensuring the stability of disturbed targets. Introducing the prior distribution by FBS, we give different object-wise weights to limit the magnitude of perturbations in the foreground and background. Besides, \cref{fig:fore} shows the adaptive \textbf{controllability} of LGP. Specifically, we set different scales $\delta$ of foregrounds to control the spaces of perturbations and update the limited regions adaptively for optimizing each object. To sum up, we intend to control perturbations in an object-wise way, not the image-level clipping in existing works\cite{DAG,CWA}. Experimental results in \cref{table:baseline} and \cref{table:LGP in different detectors.} show the excellent imperceptibility of perturbations generated by LGP. We visualize the controllable perturbations in \cref{fig:fore,fig:pic/1/per.pdf} and supplementary materials.

\begin{table}[!htbp]
\centering
\caption{\textbf{Comparisions with different budgets of perturbations.}\label{table:tradeoff}}
\resizebox{0.47\textwidth}{!}{
\begin{tabular}{cccccccc}
\hline \hline 
 & mAP$_{50}$          & PGD& PGD & PGD & PGD & LGP & LGP$_{150}$\\  \hline
\multirow{3}{*}{FR \cite{faster-rcnn}} & $\epsilon$           & 2            & 3             & 4            & 8 &   &              \\ 
& PSNR-B $\uparrow$ &  42.0           & 39.4            & 37.6            & 34.4            & 40.7        & \textbf{43.6}       \\ 
& mAP$_{50}$ $\downarrow$ &  14.3         & 8.0           & 5.0             & 2.4            & \textbf{1.5}  & \textbf{1.5} \\           \hline
\multirow{3}{*}{OR \cite{OrientedRCNN}}  &
$\epsilon$           & 1            & 2             & 4            & 8 &           &      \\ 
& PSNR-B $\uparrow$ & 47.6           & 43.9           & 38.5            & 35.2            & 47.3            & 50.6   \\ 
& mAP$_{50}$ $\downarrow$ & 50.7         & 30.3           & 15.1             & 10.0            & 4.0  & 5.2 \\ \hline \hline

\end{tabular}}

\end{table}

The \textbf{tradeoff} between attack strength and imperceptibility: In  \cref{table:tradeoff}, we use PGD$_{reg}$ to attack Faster R-CNN (FR) and Oriented R-CNN (OR) with different budgets of perturbations. As shown, the strength of the attack is inversely proportional to the imperceptibility (\ie, the higher PSNR-B is correspond with the higher mAP$_{50}$). With the same imperceptibility, LGP is more powerful than PGD. In other words, LGP shows a generic comparable attack capacity, because LGP has both the best attack strength and imperceptibility compared with baselines in \cref{table:baseline}.
\subsection{Transferability} 

In this section, we use AEs generated from the substitutive models to attack other models, which is usually called transfer-based\cite{wu2021improving,wang2021enhancing} black-box attack. 

\begin{table}[!htbp]
\centering
\caption{\textbf{Black-box against different detectors.} * are query-based attack and extracted from \cite{larget-scale}. The first column is the methods used to generate AEs, and the first row is the models to evaluate. $^{\dagger}$ denotes we combine the perturbations generated by R50, R101, and X101 against FR. We highlight the top two results in \textcolor{red}{red} and \textcolor{blue}{blue} respectively.\label{tab:Transferable baselines}}
\resizebox{0.47\textwidth}{!}{
\begin{tabular}{cccccc}
\hline \hline
    mAP      & ATSS(R101)$\downarrow$ & FCOS(X101)$\downarrow$ & GFL(X101)$\downarrow$ & DetectoRS(R101)$\downarrow$ \\  \hline
Clean$^{*}$            & 54.0            & 54.0             & 59.0            & 61.0                 \\ 
SH$^{*}$\cite{SH}            & 40.0           & 27.0            & 43.0            & 51.0                 \\ 
SQ$^{*}$\cite{SQ}             & 23.0           & 21.0             & 33.0            & 45.0                 \\ 
PRFA$^{*}$\cite{PRFA}          & 20.0            & 23.0             & 31.0            & 41.0                 \\ 
GARSDC$^{*}$\cite{larget-scale}         & \textcolor{blue}{4.0}            & 15.0             & \textcolor{blue}{16.0}            & \textcolor{blue}{28.0}                 \\ 
\hline
PGD$_{cls}$ \cite{PGD}      & 29.5           & 32.2            & 43.7           & 44.4 \\
PGD$_{reg}$ \cite{PGD}     & 17.8           & 22.6            & 43.1           & 41.7 \\
CWA  \cite{CWA}     & 28.2           & 19.6            & 44.1           & 45.6 \\
TOG  \cite{TOG}     & 20.0           & 27.5            & 42.7           & 41.4 \\
DAG \cite{DAG}      & 11.8           & \textcolor{blue}{10.6}            & 32.7           & 27.6                \\
\textbf{LGP(ours)}    & 10.1           & 10.9            & 34.2           & 30.5                  \\
\textbf{LGP$^{\dagger}$(ours)}    & \textcolor{red}{3.8}           & \textcolor{red}{8.8}            & \textcolor{red}{15.6}           & \textcolor{red}{17.5}                  \\ 
 \hline \hline
\end{tabular}}
\vspace{0.5cm}
\caption{\textbf{LGP attacks different detectors.} We use AEs generated from attacking the first column to test the mAP$_{50}$ of the first row in MS-COCO (left) and  DOTA-v1.0 (right).
\label{table:transferbaility}}
\resizebox{0.47\textwidth}{!}{
\begin{tabular}{cccc|cccc}
\hline \hline
From $\backslash$ to   & FR$\downarrow$ & TOOD$\downarrow$  & D.DETR$\downarrow$ & From $\backslash$ to           & OR$\downarrow$ & S$^{2}$A-Net$\downarrow$  & AD$\downarrow$ \\ \hline
Clean       & 51   & 51.8    & 60.7   & Clean               & 83.3   & 81.2 & 85.0           \\
FR($\gamma_1$)  & \textbf{1.5}  & 13.7   & 25.8   & OR($\gamma_1$)          & \textbf{4.0}  &24.7 & 32.3          \\
TOOD($\gamma_2$) & 17.8   & 5.9   & 24.1   & S$^{2}$A-Net($\gamma_4$)            & 38.7  & 5.2 & 37.0           \\
D.DETR($\gamma_3$) & 12.3  & 39.4   & 38.6   & AD($\gamma_5$) & 51.5  & 49.5 & 8.50           \\
$\gamma_2+\gamma_1$       & 1.90  & \textbf{0.8}   & 8.90   & $\gamma_4+\gamma_1$               & 38.7  & \textbf{5.2} & 37.0           \\
$\gamma_3+\gamma_1$       & 3.40  & 12.9   & \textbf{2.60}   & $\gamma_5+\gamma_1$               & 3.40  & 12.9 & \textbf{2.60}           \\ \hline \hline
\end{tabular}}
\end{table}

\textbf{Comparision:}
For a fair comparison, we use the same clean images and victim models in \cite{PRFA,larget-scale} to compare the mAP with query-based black-box attacks (the above of \cref{tab:Transferable baselines}). Due to cross-backbone transferability having been explored widely in classification\cite{logit,TAP,ali,byun2022improving}, we mainly focus on cross-detector transferability like prior works. Besides, we have proved that different imperceptibility has different attack strength in \cref{table:tradeoff}. So we set PSNR-B as about 40 for all transferable attacks (\ie, the bottom of \cref{tab:Transferable baselines} with budget 8) and use Faster R-CNN as the substitutive model for a fair comparison. 

In the bottom of \cref{tab:Transferable baselines}, LGP has better transferability than most baselines when they have similar imperceptibility. This is thanks to our three balanced task-oriented losses.  DAG is an untargeted classification attack, so it is more transferable than our targeted attack (\ie, Hiding Attack) to some extent. Moreover, GARSDC performs better in attacking ATSS\cite{Atss}, GFL\cite{GFL}, and DetectorRS\cite{detectors} than LGP.  But transfer-based attacks (\eg, LGP) usually are much faster than query-based attacks (\eg, GARSDC). And query-based attacks have visible perturbations which are unfair for comparisons (\eg, \cref{table:tradeoff} shows the more visible perturbations, the easier attack will be). In the last row of \cref{tab:Transferable baselines}, we combine three different backbones (\ie, R50, R101, X101) to evaluate the cross-detector transferability. Surprisingly, LGP$^{\dagger}$ gets the best transferable results which indicate we may attack any detectors by combining the attacks of one substitutive detector with multiple classical backbones.

\textbf{Transferability cross detectors:}  In \cref{table:transferbaility}\footnote{All results can be found in our supplementary materials.}, LGP makes a significant accuracy drop (decreasing about 60\% mAP$_{50}$). We have three observations from \cref{table:transferbaility}. First, AEs generated by  CNN-based and Transformer-based detectors have a large margin, indicating different types of ODs have a huge difference in their decision spaces. Even so, LGP performs a generic white-box generalization in \cref{table:LGP in different detectors.}. Secondly, AEs generated by two-stage detectors (they always have higher quality candidates) have better transferability. In other words, enough high-quality proposals play an important role in an attack. Thirdly, the value of FID is almost proportional to the transferability, which indicates imperceptible perturbations tend to lead to bad transferability. But LGP outperforms others in both aspects, indicating its strong capacity. Totally, you can get more transferable attacks by studying more generic attacks without the influence of ODs' architecture.

\textbf{Orthogonality of heterogeneous perturbations:} The above three phenomena motivate us to combine heterogeneous perturbations for better attack strength\cite{DAG}. Specifically, we can launch a new attack based on AEs generated by the other attack, and then the new AEs are the combination of the two attacks. We can effectively attack other detectors by simultaneously adding perturbations generated by typical ODs or backbones. \cref{table:transferbaility} demonstrates that attack is more powerful when we use the AEs generated by two models (\ie, $\gamma_3 + \gamma_1$ get better performance in most other detectors compared with $\gamma_1$ or $\gamma_3$). In other words, we can attack all detectors by attacking some typical detectors (\eg, CNN- and Transformer-based ODs).

\subsection{Abaltion Study}

  In this section, we do ablation studies to analyze some main choices of our proposed LGP in \cref{table:ablation study} and \cref{fig:fore_scale}. 

\textbf{The composition of attacking loss function $\cal{L}$.} With the addition of semantic, shape, and localization tasks in No.2-4, the mAP$_{50}$ values drop from 10.2 to 4.4 to 2.4 in FR, indicating balanced multi-branch attacks are stronger than single-branch. Besides, we argue that the optimization of different tasks can be guided in the same direction, by setting a high-level objective (\ie, Hiding Attack). In supplementary materials, we visualize their gradients using t-SNE\cite{TSNE}, which also shows LGP decreases the conflict among heterogeneous losses compared with RAP\cite{RAP}.

\textbf{The design of imperceptibility loss function $\cal{D}$.} We use the image-level distance constraints in No.2-5, but they are so strict that we could not get a better-attacking result. Motivated by ``deep object detectors have to look
at objects (or ROIs) to make decisions", we use FBS to encourage perturbations to attach to foregrounds in No.6. This decreases  mAP$_{50}$ by 1.0 in FR and 5.8 in OR. For flexible optimization, we update the limited regions adaptively by Adaptor in No.7. This contributes a bottleneck-breaking strength for our attack.

\textbf{The influence of different original targets ${\cal{T}}_{org}$.}
We use pre-NMS clusters $\mathbf{B}^{pre}$ in No.8 (the number is about 2000 in RPN-based detectors) and after-NMS predicts in No.9 (the number is similar to ground truth) as original targets to evaluate corresponding results. There are lots of low-quality proposals that are randomly distributed using $\mathbf{B}^{pre}$, resulting in redundant perturbations and suboptimal attack strength (uncertainty in \cref{sec:target generating}). Besides, the number of after-NMS predictions is inadequate to launch an efficient attack because different adversarial examples have different detections in different iterations (instability in \cref{sec:target generating}). The above results indicate that sufficiently stable and high-quality original targets are crucial.
\begin{figure}[!t]
  \centering
  \includegraphics[width=\linewidth]{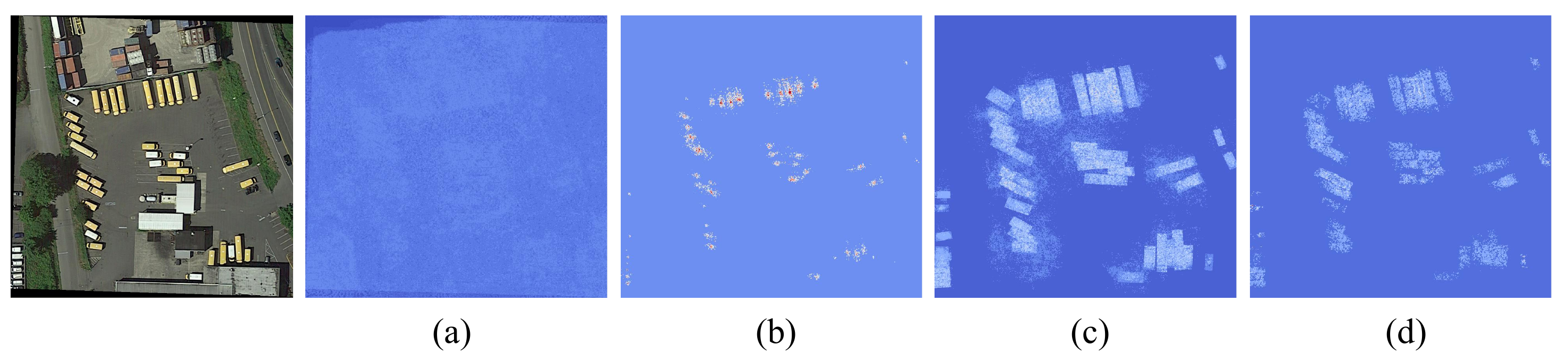}
  \vspace{-0.5cm}
  \caption{Ablation studies of controllability against Oriented R-CNN\cite{OrientedRCNN}. (a) No Limiter, (b) No Adaptor, (c) No Assigner, and (d) LGP.\label{fig:controllabilityAS}}
\end{figure}

\textbf{The visualization of controllability in different settings.} (a)We replace the Limiter with gradient-based clipping like TOG \cite{TOG}, so the image-level perturbations are generated. (b) Without the Adaptor, LGP gets perturbations with Gaussian distribution, but it cannot generate personalized perturbations according to each object itself. (c) Although the object detector gives its attention to objects, there are some perturbations out of objects after replacing HQ with pre-NMS outputs (\ie No.8 in \cref{table:ablation study})  (d) LGP uses proposal mappings to guide the ODs' attention to objects, limits perturbations with object-wise splits, and updates adaptively perturbations for a better trade-off between attack strength and imperceptibility.

\subsection{Discussions}
In this section, we further analyze the influence of different parameters on the final results. 





\textbf{More powerful.} LGP has a weak attack for some detectors, such as Gliding Vertex \cite{xu2020gliding}, RoI Transformer\cite{RoITransformer}, and ReDet\cite{ReDet}. To strengthen LGP, there are three operations make a great help. Firstly, LGP assigns more high-quality proposals as original targets for more powerful attacks in the Assigner. For example, we set $\mathbf{N}_{i}$ as 5, 25, and 50 against RoI Transformer, and the mAP$_{50}$ is 20.8, 16.0, and 9.0 respectively. Secondly, LGP makes $\lambda_{1}$ bigger in \cref{eq:overall} obviously helps more powerful strength. For example, we set N as 1 in \cref{eq:Attacker}, and the mAP$_{50}$ is 0.00 with PSNR-B 39.0 against Faster R-CNN. Thirdly, the bigger perturbed spaces mean better attack strength. LGP gets 1.1 mAP$_{50}$ and 42.3 PSNR-B after replacing the limiter with gradients clipping like TOG\cite{TOG}. 

\textbf{More controllable.} In \cref{fig:fore_scale}, the bigger scale of foregrounds, the lower mAP$_{50}$ and PSNR-B.  This comparison verifies that stronger attack capacity will always come at the expense of bigger space for perturbations and lower image quality. Due to the final perturbations being learnable, we can control the distribution of perturbations according to practical requirements (\eg, LGP with the value 0.5 of $\delta$ also has a comparable result). Besides, we use simple Gaussian distribution to weight adversarial perturbations, but other distributions also can be applied to guide the optimization (\eg, a prior patch like \cite{liu2018dpatch}).

\textbf{Limitations.} Due to the generic ability against different detectors, LGP always has slower speeds for constructing Adversarial Examples than other methods which leverage special structures of object detectors in one-stage detectors. Specifically, LGP needs to select, assign, and split attacked targets from thousands of candidates, but others filter low-quality proposals with a threshold.

\textbf{Future works.} LGP has three key and imperfect modules, \ie, the Assigner, Attacker, and Limiter. Whether a quicker assign strategy  could be designed? Whether other types of attacks could get more powerful results? For example, untargeted attacks. Whether other weights could get more powerful results with smaller perturbed spaces? In other words, LGP may get a powerful strength with some imperceptible patches attached to objects which induces an object-wise imperceptible physical attack like\cite{patch}.

\section{Conclusion}
In this paper, our main purposes are not to design a more powerful and imperceptible white-box attack. Motivated by the unique behaviors of object detectors, we formulate the adversarial attack against object detection as a detector- and dataset-agnostic, and object-wise optimization problem. Hence generic and controllable LGP is designed against object detection. Unlike the existing attack methods that fool detector-intrinsic structures with image-level perturbations, LGP only considers a small part of detectors' outputs to optimize jointly multi-task gradients and object-wise controllable constraints. Comprehensive experiments across most advanced detectors show that LGP can yield adversarial examples with controllable perturbations without leveraging any specific structures of detectors.

{\small
\bibliographystyle{ieee_fullname}
\bibliography{egbib}
}

\end{document}